\pdfoutput=1

\documentclass[11pt]{article}

\usepackage[final]{EMNLP2022}

\usepackage{times}
\usepackage{latexsym}

\usepackage{hyperref}       
\usepackage{url}            
\usepackage{booktabs}       
\usepackage{amsfonts}       
\usepackage{nicefrac}       
\usepackage{wrapfig,lipsum}
\usepackage{amsmath}
\usepackage{algorithm}
\usepackage[noend]{algpseudocode}
\usepackage{multirow}
\usepackage{arydshln}
\usepackage{subcaption}
\usepackage{pgfplots}
\usepackage{tikz}
\usepackage{mathtools}
\usepackage{xfrac}

\usepackage[T1]{fontenc}

\usepackage[utf8]{inputenc}

\usepackage{microtype}

\usepackage{inconsolata}

\newcommand{\fedweitFullName}{Federated Weighted Inter-client Transfer}
\newcommand{\fedweitAcronym}{FedWeIT}

\newcommand{\proposedModelFullName}{Federated Selective Inter-client Transfer}
\newcommand{\proposedModelAcronym}{FedSeIT}

\newcommand{\SITFullName}{Selective Inter-client Transfer}
\newcommand{\SITAcronym}{SIT}

\DeclareMathOperator{\thetaparam}{\boldsymbol{\theta}}
\newcommand{\B}{\ensuremath{\mathbf{B}}}
\newcommand{\A}{\ensuremath{\mathbf{A}}}
\newcommand{\m}{\ensuremath{\mathbf{m}}}
\newcommand{\matrixparamdim}{\ensuremath{\{\mathbb{R}^{\mathcal{I}_l \times \mathcal{O}_l}\}_{l=1}^{L_{FC}}}}
\newcommand{\filterparamdim}{\ensuremath{\{\mathbb{R}^{\mathcal{F}_l \times \mathcal{D} \times \mathcal{N}_l^\mathcal{F}}\}_{l=1}^{L_{Conv}}}}
\newcommand{\filterparamdimreduced}{\ensuremath{\mathbb{R}^{\mathcal{F} \times \mathcal{D} \times \mathcal{N}^\mathcal{F}}}}

\newcommand{\server}{\ensuremath{s}}
\newcommand{\interference}{interference}
\newcommand{\icinterference}{\textit{inter-client interference}}
\newcommand{\homogeneous}{homogeneous}
\newcommand{\heterogeneous}{heterogeneous}
\newcommand{\cl}{\textit{continual learning}}
\newcommand{\fl}{\textit{federated learning}}
\newcommand{\fcl}{\textit{federated continual learning}}
\newcommand{\wspace}{ }
\newcommand{\CNN}{\ensuremath{\mathtt{CNN}}}
\newcommand{\softmax}{\ensuremath{\mathtt{Softmax}}}
\newcommand{\withspace}[1]{#1\;}
\newcommand{\X}{\ensuremath{\mathbf{X}}}
\newcommand{\z}{\ensuremath{\mathbf{z}}}
\newcommand{\W}{\ensuremath{\mathbf{W}}}
\newcommand{\tablepm}[1]{{\scriptsize $\pm$ #1}}
\newcommand{\dataset}{\mathbb{D}}
\DeclareMathOperator*{\minimize}{minimize\;}

\title{
	Federated Continual Learning for Text Classification \\via Selective Inter-client Transfer}

\newcommand*{\affaddr}[1]{#1} 
\newcommand*{\affmark}[1][*]{\textsuperscript{#1}}

\author{Yatin Chaudhary\affmark[1,2],  Pranav Rai\affmark[1,2], 
	Matthias Schubert\affmark[2], Hinrich Sch\"{u}tze\affmark[2], Pankaj Gupta\affmark[1] \\ 
	\affaddr{
		{\affmark[1]DRIMCo GmbH, Munich, Germany} $|$ 
		{\affmark[2]University of Munich (LMU), Munich, Germany}}\\
	{\tt \{{firstname.lastname\}}@drimco.net}
} 

\begin{document}
	\maketitle
	\begin{abstract}
		In this work, we combine the two paradigms: Federated Learning (FL) and Continual Learning (CL) for text classification task in cloud-edge continuum. 
		The objective of Federated Continual Learning (FCL) is to improve deep learning models over life time at each client by (relevant and efficient) knowledge transfer without sharing data. 
		Here, we address challenges in minimizing inter-client interference while knowledge sharing due to heterogeneous tasks across clients in FCL setup.
		In doing so, we propose a novel framework, \textit{\proposedModelFullName\wspace} (\proposedModelAcronym) which selectively combines model parameters of foreign clients.
		To further maximize knowledge transfer, we assess domain overlap and select informative tasks from the sequence of historical tasks at each foreign client while preserving privacy.
		Evaluating against the baselines, we  show improved performance, a gain of (average) 12.4\% in text classification over a sequence of tasks using five datasets from diverse domains. To the best of our knowledge, this is the first work that applies FCL to NLP.
	\end{abstract}
	
	\section{Introduction}

	\textbf{Federated Learning} \cite{yurochkin2019bayesian,li2020federated,zhang2020federated,karimireddy2020scaffold,caldas2018federated} in Edge Computing\footnote{extends cloud computing services closer to data sources} \cite{edgecomputing} has gain attraction in recent years due to (a) data privacy and sovereignty- especially imposed by government regulations (GDPR, CCPA etc.), and (b) the need for sharing knowledge across edge (client) devices such as mobile phones, automobiles, wearable gadgets, etc. while maintaining data localization. Federated Learning (FL) is a privacy-preserving machine learning (ML) technique that enables collaborative training of ML models by sharing model parameters across distributed clients through a central server - without sharing their data. In doing so, a central server aggregates model parameters from each participating client and then distribute the aggregated parameters, where ML models at each client are optimized using them - achieving inter-client transfer learning. In this direction, the recent works such as FedAvg~\cite{fedavg}, FedProx~\cite{li2020federated}, FedCurv~\cite{fedcurv} have introduced parameter aggregation techniques and shown improved learning at local clients  - augmented by the parameters of foreign clients.

	On the other hand, the edge devices generate a continuous stream of data where the data distribution can drift over time; therefore, the need for Continual Learning like humans do. \textbf{Continual Learning} (CL)~\cite{thrun1995lifelong,kumar2012learning,kirkpatrick2017overcoming,schwarz2018progress,gupta2020neural} empowers deep learning models to continually accumulate knowledge from a sequence of tasks  - reusing historical knowledge while minimizing catastrophic forgetting (drift in learning of the historical tasks) over life time.
	
	
	\textbf{Federated Continual Learning (FCL):}
	This work investigates the combination of the two paradigms of ML: Federated Learning and Continual Learning with an objective to model a sequence of tasks over time at each client via inter-client transfer learning while preserving privacy and addressing heterogeneity of tasks across clients.  
	There are two key challenges of FCL: 
	(1) \textit{catastrophic forgetting}, and  
	(2) \textit{inter-client interference} due to heterogeneity of tasks (domains) at clients. At central server, FedAvg \cite{fedavg} aggregates-averages model parameters from each client without considering inter-client interference. 
	To address this, \fedweitAcronym\wspace \cite{DBLP:conf/icml/YoonJLYH21} approach performs FCL by sharing \textit{task-generic} (via dense base parameters) and \textit{task-specific} (via task-adaptive parameters) knowledge across clients. In doing so, at the server, they aggregate the dense base parameters however, no aggregation of the task-adaptive parameters, and then broadcast both the types of parameters. See further details in Figure \ref{fig:client_server_communication} and section \ref{sec:proposedmodel}.
	\fedweitAcronym, the first approach in FCL, investigates computer vision tasks (e.g., image classification), however the technique has limitations in aligning domains of foreign clients while augmented learning  at each local client using task-adaptive parameters - that are often misaligned with local model parameters 
	in parameter space~\cite{fedavg} due to heterogeneity in tasks. Therefore, a simple weighted additive composition technique does not address inter-client interference and determine domain relevance in foreign clients while performing transfer learning.
	
	\textbf{Contributions}: To the best of our knowledge, this is the first work that applies FCL to NLP task (text classification). At each local client, to maximize the inter-client transfer learning and minimize inter-client interference,  we propose a novel approach, \textit{\proposedModelFullName}\wspace (\proposedModelAcronym) that aligns domains of the foreign task-adaptive parameters via projection in the augmented transfer learning.  
	To exploit the effectiveness of domain-relevance in handling a number of foreign clients, we further extend \proposedModelAcronym\wspace by a novel task selection strategy, \textit{\SITFullName\wspace} (\SITAcronym) that efficiently selects the relevant task-adaptive parameters from the historical tasks of (many) foreign clients - assessing domain overlap at the global server using encoded data representations while preserving privacy.
	We evaluate our proposed approaches: \proposedModelAcronym\wspace and \SITAcronym\wspace for Text Classification task in FCL setup using five NLP datasets from diverse domains and show that they outperforms existing methods.
	Our main contributions are as follows:
	
	\textbf{(1)} We have introduced Federated Continual Learning paradigm to NLP task of text classification that collaboratively learns deep learning models at distributed clients through a global server while maintaining data localisation and continually learn over a sequence of tasks over life time - minimizing catastrophic forgetting, minimizing inter-client interference and maximizing inter-client knowledge transfer. 
	
	\textbf{(2)} We have presented novel techniques:  \proposedModelAcronym\wspace and \SITAcronym\wspace that align domains and select relevant task-adaptive parameters of the foreign clients while augmented transfer learning at each client via a global server. Evaluating against the baselines, we have demonstrated improved performance, a gain of (average) 12.4\% in text classification over a sequence of tasks using 5 datasets. Implementation of \proposedModelAcronym\wspace is available at \url{https://github.com/RaiPranav/FCL-FedSeIT} (See appendix \ref{sec:exp_settings_appendix} and \ref{sec:reproducibility}).

	\section{Methodology}
	
	\begin{table}[t]
		\center
		\resizebox{.495\textwidth}{!}{
			\begin{tabular}{c|l}
				\multicolumn{1}{c|}{\bf Notation} & \multicolumn{1}{c}{\bf Description} \\ 
				\midrule
				$c$, $s$ & current client, global server \\
				$t$, $r$ & current task, current round \\
				$C$, $T$, $R$ & total number of clients, tasks, rounds \\
				$\mathcal{T}_c^t$ & training dataset for task $t$ of client $c$ \\
				$\thetaparam_c^t$ & model parameter set for task $t$ of client $c$ \\
				$\thetaparam_G$ & global aggregated server parameter \\
				$x_i^t, y_i^t$ & input document, label pair in $\mathcal{T}_c^t$ \\
				$\z, \hat{y}$ & \CNN\wspace output dense vector; predicted label \\
				$\B_c^t, \A_c^t$ & local base and task-adaptive parameters \\
				$\alpha_c^t, \m_c^t$ & scalar attention parameters, mask parameters \\
				$\oplus$ & concatenation operation \\
				$\W_c^t, \W_f^t$ & projection matrices: alignment and augmentation \\
				$K$ & number of parameters selected for transfer in \SITAcronym \\
				$\mathcal{D}, \dataset$ & word embedding dimension, dataset \\
				$L_d, L_c^t$ & unique labels in dataset $d$, each task dataset $\mathcal{T}_c^t$ \\
				$\mathcal{F}$ & filter size of convolution layer \\
				$\mathcal{N}^\mathcal{F}$ & number of filters in convolution layer \\
				$\lambda_1, \lambda_2$ & hyperparameters: sparsity, catastrophic forgetting \\
		\end{tabular}}
		\caption{Description of the notations used in this work where matrices and vectors are denoted by uppercase and lowercase bold characters respectively.
		}
		\label{tab:notations}
	\end{table}

	\begin{figure*}[h]
		\centering
		\includegraphics[width=\textwidth]{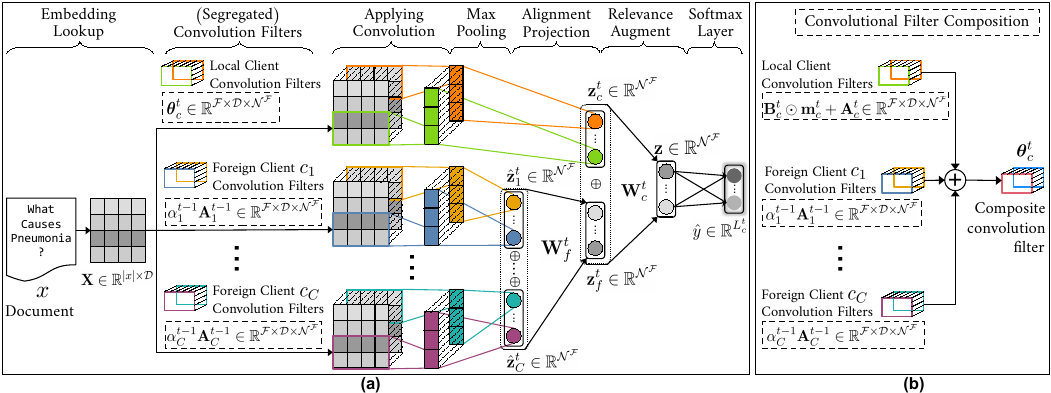}
		\caption{(a)  Illustration of the proposed \proposedModelAcronym\wspace framework where, task-adaptive parameters of foreign clients are segregated and domain-aligned for selective utilization. \textit{HowToRead}: Note the coloring scheme in convolution filters of local and foreign clients and their application in convolution. (b) Weighted additive filter composition performed in the baseline model: \fedweitAcronym. Note the composite  $\thetaparam_c^t$ vs segregated convolution filters of \proposedModelAcronym.}  
		\label{fig:fedseit}
	\end{figure*}
	
	\subsection{Federated Continual Learning}
	
	Consider a global server \server\wspace and $C$ distributed clients, such that each client $c_c \in \{c_1, . . . , c_C\}$ learns a local ML model on its privately accessible sequence of tasks $\{1,...,t,...,T\}$ with datasets $\mathcal{T}_c \equiv  \{\mathcal{T}_c^1,...,\mathcal{T}_c^t,...,\mathcal{T}_c^T\}$, where $\mathcal{T}_c^t = \{x_i^t, y_i^t\}_{i=1}^{N^t}$ is a labeled dataset for $t_{th}$ task consisting of $N^t$ pairs of documents $x_i^t$ and their corresponding labels $y_i^t$. 
	Please note that there is no relation among the datasets $\{\mathcal{T}_{c}^t\}_{c=1}^C$ for task $t$ across all clients. 
	Now, in each training round $r \in \{1,...,R\}$ for task $t$, the training within FCL setup can be broken down into three steps: 
	
	\textbf{Continual learning at client:} Each client $c_c$ effectively optimizes its model parameters $\thetaparam_c^{t(r)}$ (for task $t$) using task dataset $\mathcal{T}_c^t$ in a continual learning setting such that: 
	(a) it minimizes \textit{catastrophic forgetting} of past tasks, and 
	(b) it boosts learning on the current task using the knowledge accumulated from the past tasks.
	
	\textbf{Parameter aggregation at server:} After training on task $t$, each client $c_c$ transmits updated model parameters $\thetaparam_c^{t(r)}$ to the server \server\wspace and server aggregates them into the global parameter $\thetaparam_G$ to accumulate the knowledge across all clients. 
	
	\textbf{Inter-client knowledge transfer:} Server transmits global aggregated parameter $\thetaparam_G$ to all participating clients for inter-client knowledge transfer in the next training round $r+1$.

	\textbf{Challenges:} 
	However, there are two main sources of \icinterference\wspace within FCL setup: 
	(1) using a single global parameter $\thetaparam_G$ during parameter aggregation at server to capture the cross-client knowledge~\cite{DBLP:conf/icml/YoonJLYH21} due to model parameters trained on irrelevant foreign client tasks, and
	(2) \textit{non-alignment} of the foreign client model parameters given the heterogeneous task domains across clients.
	This leads to the hindrance of the local model training at each client by updating its parameters in erroneous directions, thus resulting in: (a) \textit{catastrophic forgetting} of the client's historical tasks, and (b) sub-optimal learning of client's current task.
	For brevity, we will omit notation of round $r$ from further equations and mathematical formulation except algorithms.

	\subsection{\proposedModelFullName}\label{sec:proposedmodel}

	To tackle the above-mentioned challenges, we propose \proposedModelFullName\wspace(\proposedModelAcronym) framework which aims to minimize \icinterference\wspace and communication cost while maximizing inter-client knowledge transfer in FCL paradigm.
	Motivated by~\citet{DBLP:conf/icml/YoonJLYH21}, \proposedModelAcronym\wspace model decomposes each client's model parameters $\thetaparam_c^t$ into a set of three different parameters: 
	(1) dense \textit{local base parameters} $\B_c^t$ which captures and accumulates the \textit{task-generic knowledge} across client's private task sequence $\mathcal{T}_c$, 
	(2) sparse \textit{task-adaptive parameters} $\A_c^t$ which captures the \textit{task-specific knowledge} for each task in $\mathcal{T}_c$, and 
	(3) sparse \textit{mask parameters} $\m_c^t$ which allow client model to selectively utilize the global knowledge. 
	For each client $c_c$, $\B_c^t$  is randomly initialized only once before training on the first task and shared throughout the task sequence $\mathcal{T}_c$, 
	while a new $\A_c^t$ and $\m_c^t$ parameters are initialized for each task $t$.
	At the global server, we have global parameter $\thetaparam_G$ which accumulates task-generic knowledge across all clients i.e., global knowledge, by aggregating local base parameters sent from all clients. 
	Finally, for each client $c_c$ and task $t$, the model parameters $\thetaparam_c^t$ can be described as:
	\begin{equation}\label{eq:theta_decomposition}
		\begin{aligned}
			\B_c^t &\leftarrow \thetaparam_G \\
			\thetaparam_c^t &= \B_c^t \odot \m_c^t + \A_c^t \\
		\end{aligned}
	\end{equation}
	where, each client initializes $\B_c^t$ using $\thetaparam_G$ received from the server containing global knowledge, before training on task $t$, to enable inter-client knowledge transfer.
	Therefore, the first term signifies selective utilization of global knowledge using the mask parameter $\m_c^t$, which restricts the impact of \icinterference\wspace during server aggregation.
	Due to additive decomposition of parameters, the second term $\A_c^t$ captures task specific knowledge.
	
\begin{figure}[]
		\centering
		\includegraphics[width=0.48\textwidth]{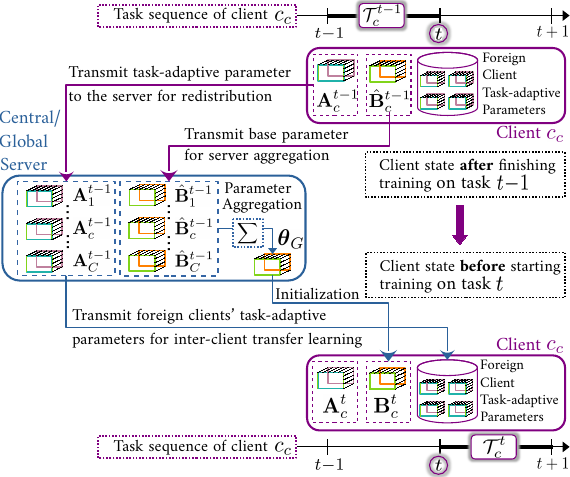}
		\caption{Inter-client Transfer Learning in FCL: (top) broadcasting client-to-server model parameters of client  $c_c$ after training the task $t-1$; (middle) parameter aggregation at the server;  and (bottom) reception of the parameters at the client $c_c$ before training the task $t$}
	\label{fig:client_server_communication}
\end{figure}

	Another key benefit of parameter decomposition is that by accessing task-adaptive parameters $\A_c^t$ of the past tasks from foreign clients, a client can selectively utilize task-specific knowledge of the relevant tasks, thus further minimizing \icinterference\wspace and maximizing knowledge transfer.
	Therefore, before training on task $t$ in \proposedModelAcronym, global server enables distribution of task-adaptive parameters of the past tasks from all clients i.e, $\{\A_c^{t-1}\}_{c=1}^C$, to each client.
	However, as already discussed earlier, given the \heterogeneous\wspace task domains across clients, the parameters learned during training at foreign clients are often \textit{non-aligned} in the parameter space.
	So, \proposedModelAcronym\wspace addresses this issue by treating each foreign task-adaptive parameter separately and align their local model output vectors via projection to augment the relevant knowledge during learning on the current task. 
	Also, for each foreign task-adaptive parameter, we introduce an attention parameter $\alpha$ to further control the impact of \icinterference.
	Figure~\ref{fig:client_server_communication} illustrates the parameter transmissions between each client and the server before training on task $t$.
	
\textbf{Client activity:} Before training on task $t$, each client $c_c$ receives the global parameter $\thetaparam_G$ and $C$ foreign clients' task-adaptive parameters  $\{\A_c^{t-1}\}_{c=1}^C$ from the server.
Then, each client partially updates its base parameter $\B_c^t$ with the non-zero entries of $\thetaparam_G$ i.e., $\B_c^t(n) = \thetaparam_G(n)$ where $n$ is the nonzero element of $\thetaparam_G$. 
After optimizing local model parameters for task $t$ in the continual learning setup, the sparsified base parameter $\hat{\B}_c^t = \B_c^t \odot \m_c^t$ and the task-adaptive parameter $\A_c^t$ are transmitted to the global server \server.

\textbf{Server activity:} After training on task $t$, the global server receives local base parameters $\{\hat{\B}_c^t\}_{c=1}^C$ and task-adaptive parameters $\{\A_c^t\}_{c=1}^C$ from all clients. 
To capture global knowledge, the server aggregates base parameters $\{\hat{\B}_c^t\}_{c=1}^C$ into the global parameter $\thetaparam_G$.
Then, before training on the next task $t+1$, server transmits the task-adaptive parameters $\{\A_c^t\}_{c=1}^C$ along with the global parameter $\thetaparam_G$ to each client for inter-client knowledge transfer via selective utilization.

\label{sec:client_model_appendix}
\begin{figure}[]
	\centering
	\includegraphics[scale=0.99]{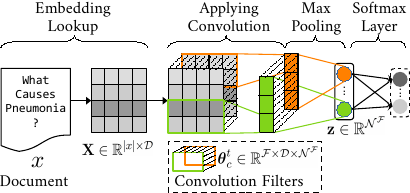}
	\caption{A detailed illustration of CNN client model used in \proposedModelAcronym\wspace framework.}
	\label{fig:cnn}
\end{figure}

\textbf{Method:} 
In this work, we use Convolutional Neural Network (CNN)~\cite{kim-2014-convolutional}, illustrated in Figure~\ref{fig:cnn}, as the \cl\wspace client model in \fcl\wspace setting. 
So, the client model parameters $\B_c^t, \A_c^t, \thetaparam_c^t \in \filterparamdimreduced$ are a set of convolutional filters where, $\mathcal{F}$ is the filter size, $\mathcal{D}$ is the word embedding dimension and $\mathcal{N}^\mathcal{F}$ is the count of filters.
See appendix~\ref{sec:client_model_appendix} for detailed model description.

Consider training for task $t$ at client $c_c$ using dataset $\mathcal{T}_c^t \equiv \{x_i^t, y_i^t\}_{i=1}^{N^t}$, the \proposedModelAcronym\wspace framework segregates client's local model parameters $\thetaparam_c^t$ from the foreign client's task-adaptive parameters $\{\A_c^{t-1}\}_{c=1}^C$. 
As illustrated in Figure~\ref{fig:fedseit}(a), for an input document $x \in \mathcal{T}_c^t$ an embedding matrix $\X \in \mathbb{R}^{|x| \times \mathcal{D}}$ is generated via embedding lookup and then using $\X$ the CNN model computes: 
(1) one client dense vector $\z_c^t \in \mathbb{R}^{\mathcal{N}^{\mathcal{F}}}$ using local client parameters $\thetaparam_c^t$, and 
(2) $C$ foreign dense vectors $\hat{\z}_i^t \in \mathbb{R}^{\mathcal{N}^{\mathcal{F}}}$ using foreign clients' task-adaptive parameters $\{\A_c^{t-1}\}_{c=1}^C$ as: 
\begin{equation}\label{eq:cnn_output_proposed_method}
	\begin{aligned}
		\z_c^t &= \CNN(x,\thetaparam_c^t) \\
		\hat{\z}_i^t &= \CNN(x,\alpha_i^{t-1}\A_i^{t-1}) \\
	\end{aligned}
\end{equation}
where, $i \in \{1,...,C\}$, $|x|$ is the count of tokens in $x$, $\alpha$ is the attention parameter and \CNN\wspace is the convolution \& max-pooling function.
Then we align the foreign parameters in parameter space by concatenating and projecting all foreign dense vectors $\{\hat{\z}_i^t\}$ to get a single foreign vector $\z_f^t \in \mathbb{R}^{\mathcal{N}^{\mathcal{F}}}$.	Finally, we selectively augment the relevant knowledge from foreign clients by concatenating and projecting $\z_c^t$, $\z_f^t$ for prediction of label $\hat{y}$ as:
\begin{equation}\label{eq:proposed_method_deep}
	\begin{aligned}
		\z_f^t &= \W_f^t \{\hat{\z}_1^t \oplus ... \oplus \hat{\z}_C^t\} \\
		\z &= \W_c^t\{\z_c^t \oplus \z_f^t\} \\
		\hat{y} &= \softmax(\z) 
	\end{aligned}
\end{equation}
where $\W_f^t \in \mathbb{R}^{C\mathcal{N}^{\mathcal{F}} \times \mathcal{N}^{\mathcal{F}}}$, $\W_c^t \in \mathbb{R}^{2\mathcal{N}^{\mathcal{F}} \times \mathcal{N}^{\mathcal{F}}}$ are the projection matrices, $\z \in \mathbb{R}^{\mathcal{N}^{\mathcal{F}}}$ is the augmented vector, $\softmax$ is a softmax layer, $\hat{y} \in \mathbb{R}^{L_c^t}$ is the model prediction and $L_c^t$ is the unique label set in $\mathcal{T}_c^t$.
By segregation of foreign parameters via concatenation and projection of the $\CNN$ outputs, \proposedModelAcronym\wspace enables local client model to align foreign parameters in feature space and augment knowledge from the relevant foreign tasks.
In doing so, \proposedModelAcronym\wspace effectively circumvents the \icinterference\wspace due to \heterogeneous\wspace task domains while maximizing inter-client transfers. 

\textbf{Comparison with \fedweitAcronym}: 
As illustrated in Figure~\ref{fig:fedseit}(b), in contrast to our approach, in the baseline \fedweitAcronym\wspace framework each client performs inter-client knowledge transfer via weighted composition of foreign task-adaptive parameters i.e., convolution filters, along with local parameters as:
\begin{equation}\label{eq:theta_decomposition_fedweit}
	\thetaparam_c^t = \B_c^t \odot \m_c^t + \A_c^t + \sum_{i \in C}\alpha_i^{t-1}\A_i^{t-1}
\end{equation} 
However, due to heterogeneity of task domains across clients, the simple additive aggregation of foreign client parameters with local parameters leads to \icinterference\wspace and sub-optimal learning of the local task.
Unlike \fedweitAcronym\wspace (where the attention $\alpha$ parameters decide the relevance of foreign client parameters), the segregation of foreign client parameters in our proposed \proposedModelAcronym\wspace method enables the local model to selective augment relevant knowledge from foreign clients' task-adaptive parameters via projection.

\textbf{Training:} 
For task $t$ at each client $c_c$, the FCL optimization objective for local client model can be decomposed into three components as:
\begin{equation}\label{eq:fedweit_obj}
	\begin{aligned}
		\minimize_{\A_c^t, \B_c^t, \m_c^t} &\sum_{i=1}^{N_t}\mathcal{L}(\hat{y}_i^t,y_i^t) \\
		&+ \lambda_1\mathcal{L}_{sp}([\m_c^t, \A_c^{1:t}]) \\
		&+ \lambda_2\sum_{i=1}^{t-1}||\Delta\B_c^t\odot\m_c^i-\Delta\A_c^i||_2^2 \\
	\end{aligned}
\end{equation}
where, $\hat{y}_i^t$ and $y_i^t$ are the predicted and true label respectively for input $x_i^t$ and $N_t$ is the number of documents.
The first term is the model training objective for current task $t$.
The second term is a sparsity objective to induce sparsity in the mask $\m_c^t$ and task specific parameters $\A_c^t$ for efficient server-client communication, where $\lambda_1$ is a hyperparameter to regulate sparsity. 
The final term is the \textit{continual learning} regularization~\cite{kirkpatrick2017overcoming} objective to minimize \textit{catastrophic forgetting} by controlling the drift in parameters learned from the past tasks.
Here, $\Delta\B_c^t$ is the change in $\B_c^t$ between the current and previous task i.e., $\Delta\B_c^t = \B_c^t-\B_c^{t-1}$, $\Delta\A_c^i$ is the change in task-adaptive parameters for task $i$ between the current and previous time-step and $\lambda_2$ is the hyperparameter to regulate sparsity.
The task specific parameters $\A_c^{1:t-1}$ of the past tasks are updated to balance the change in $\B_c^t$ i.e., $\Delta\B_c^t$, which minimizes drift in the solutions learned for the past tasks.
For higher values of $\lambda_2$, the training objective set high penalty on forgetting of the previous tasks.
Algorithm~\ref{algo:fedseit} describes the \proposedModelAcronym\wspace framework where, the server aggregation function $\mathtt{Agg}$ is agnostic to the choice of available methods like FedAvg, FedProx etc.
In \proposedModelAcronym\wspace, we use FedAvg.

\begin{algorithm}[t]
	\small
	\centering
	\caption{Proposed FedSeIT Framework}\label{algo:fedseit}
	\begin{algorithmic}[1]
		\Statex \textbf{Input}: Task datasets $\{\mathcal{T}_c^{1:t}\}_{c=1}^C$, Global parameter $\thetaparam_G$
		\Statex \textbf{Output}: $\{\B_c^t, \m_c^{1:t}, \alpha_c^{1:t}, \A_c^{1:t}, \W_c^{1:t}, \W_f^{1:t}\}_{c=1}^C$
		\State Server initializes $\thetaparam_G$
		\State Each client $c \in \mathcal{C} \equiv \{1,...,C\}$ connects with server
		\For {task $t = 1, ..., T$}
			\For {round $r = 1, ..., R$}
				\State Server transmits global parameter $\thetaparam_G$ to all $c \in \mathcal{C}$
				\State Each client $c \in \mathcal{C}$ initializes $\B_c^t$ using $\thetaparam_G$
				\If {$r = 1$}
					\State Each client $c \in \mathcal{C}$ initializes $\A_c^t$, $\m_c^t$
				\EndIf
				\If {$r = 1$ \textbf{and} $t \ne 1$}
				\State Server transmits $\{\A_c^{(t-1,R)}\}_{c=1}^C$ to all $c \in \mathcal{C}$
				\EndIf
				\State Prepare model parameters: $\thetaparam_c^t \leftarrow \B_c^t \odot \m_c^t + \A_c^t$
				\State Each client $c \in \mathcal{C}$ minimizes equation \ref{eq:fedweit_obj} using equations~\ref{eq:theta_decomposition},\ref{eq:cnn_output_proposed_method},\ref{eq:proposed_method_deep} to learn task $t$ in continual learning setup
				\State Each client $c \in \mathcal{C}$ transmits $\hat{\B}_c^{(t,r)}$ to server
				\State $\thetaparam_G \leftarrow \mathtt{Agg}(\{\hat{\B}_c^{(t,r)}\}_{c \in \mathcal{C}}) = \frac{1}{|\mathcal{C}|}\sum_{c=1}^{C}\hat{\B}_c^{(t,r)}$
				\State Server distributes $\thetaparam_G$ to all clients $c \in \mathcal{C}$
				\If {$r = R$ \textbf{and} \SITAcronym\wspace is enabled}
					\State Each client computes $\hat{\mathcal{T}_c^t}$ using equation~\ref{eq:sit}
					\State Each client transmits $\{\hat{\mathcal{T}_c^t},\A_c^{(t,R)}\}$ to server
					\State Server computes task-task similarity using $\hat{\mathcal{T}_c^t}$
					\State Server selects top-$K$ parameters $\forall c \in \mathcal{C}$
					\State Server distributes $\{\A^{(k,R)}\}_{k=1}^K$ to all $c \in \mathcal{C}$
				\EndIf
				\If {$r = R$ \textbf{and} \SITAcronym\wspace is not enabled}
					\State Each client transmits $\A_c^{(t,R)}$ to server
					\State Server distributes $\{\A_c^{(t,R)}\}_{c=1}^C$ to all $c \in \mathcal{C}$
				\EndIf
			\EndFor
		\EndFor
	\end{algorithmic}
\end{algorithm} 

\begin{figure*}[]
	\centering
	\includegraphics[width=\textwidth]{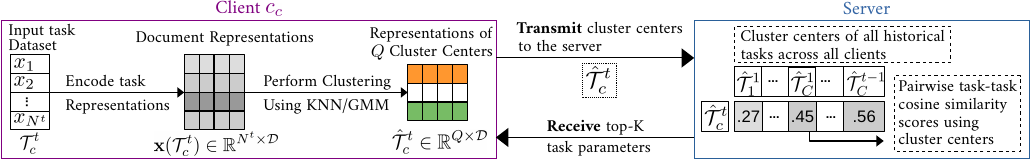}
	\caption{Detailed illustration of the proposed \SITAcronym\wspace method, where foreign task-adaptive parameters are selected based on task domain-relevance to maximize inter-client transfer learning and minimize communication cost.}
	\label{fig:sit}
\end{figure*}

\textbf{Dense Layer Sharing (DLS)}: 
Unlike the convolution filters of CNN model that captures the transferable n-gram patterns in the data, the dense layer parameters $\{\W_c^t$, $\W_f^t\}$ capture fine-grained alignment information based on the selection and ordering of unique output labels $y$ and foreign client parameters $\{\A_c^{t-1}\}_{c=1}^C$. 
In heterogeneity, the server aggregation and distribution of dense layer parameters introduce sub-optimal initialization point for training of future tasks across clients. 
Therefore, in \proposedModelAcronym, we do not share projection layer parameters with server by default.
In doing so, we also increase client privacy in response to adversarial attacks. 
To validate our hypothesis, we evaluate a variant of our proposed model \textbf{\proposedModelAcronym+DLS}, where we share the dense projection layer parameters $\{\W_c^t$, $\W_f^t\}$ in \proposedModelAcronym\wspace framework.

\subsection{\SITFullName\wspace(\SITAcronym)}
In \proposedModelAcronym, before training on task $t$, each client receives $C$ foreign task-adaptive parameters $\{\A_1^{t-1}, ...,\A_C^{t-1}\}$ of the previous task $t-1$ from each client via server \server\wspace for inter-client transfer learning.
However, given task heterogeneity, the previous task parameters might be irrelevant for learning current task which could lead to \icinterference.
To resolve this, we could transmit parameters of all historical tasks from all foreign clients i.e.,  $\{\A_1^1, ...,\A_C^{t-1}\}$, to minimize interference by finding relevant parameters.
But, this could lead to burgeoning of computational complexity and communication cost increasing with each completed task.
Therefore, to tackle this, we propose \SITFullName\wspace (\SITAcronym) method, which uses encoded task representations to efficiently explore all historical tasks across foreign clients via domain overlap, and selects the relevant parameters to minimize \icinterference\wspace and maximize knowledge transfer.

\textbf{Client:} For each task $t$, before training, each client $c_c$ generates the encoded vector representation $\mathbf{x}(x) \in \mathbb{R}^{\mathcal{D}}$ for each document $x$ in task dataset $\mathcal{T}_c^t$ via embedding lookup and averaging.
After that, using K-Nearest Neighbor (KNN) or Gaussian Mixture Model (GMM) algorithm, each client performs clustering on the encoded vector representations as follows:
\begin{equation}\label{eq:sit}
	\begin{aligned}
		\mathbf{x}(x) &= \frac{1}{|x|} \sum_{i=1}^{|x|} \mathtt{EmbeddingLookup}(x_i, \textbf{E}) \\
		\hat{\mathcal{T}}_c^t &= \mathtt{Clustering}(\{\mathbf{x}(x) | \forall x \in \mathcal{T}_c^t\}) \\
	\end{aligned}
\end{equation}
where, $\hat{\mathcal{T}}_c^t \in \mathbb{R}^{Q \times \mathcal{D}}$ are the representations of the cluster centers, $\mathtt{Clustering} \in \{\mbox{KNN}, \mbox{GMM}\}$, $Q$ denotes the number of cluster centers and $\textbf{E}$ is a pre-trained word embedding repository like Word2Vec~\cite{mikolov2013efficient}.
Ultimately, each client transmits $\hat{\mathcal{T}}_c^t$ to the global server.

\textbf{Server:} Once the server receives representations of cluster centers $\hat{\mathcal{T}}_c^t$ of task $t$ from client $c_c$, it computes pairwise task-task domain overlap using average cosine-similarity score between the current client task $\hat{\mathcal{T}}_c^t$ and each historical task across all clients $\{\hat{\mathcal{T}}_1^1, ..., \hat{\mathcal{T}}_C^{t-1}\}$.
The server then selects and transmits top-$K$ relevant (high similarity) parameters to the client for inter-client transfer learning, where $K$ is a hyperparameter.
Therefore, in scenarios where task history is long ($t > 10$) and/or clients are too many ($C > 10$), \SITAcronym\wspace can keep the computational complexity of client model constant while minimizing \icinterference\wspace and maximizing knowledge transfer.
To test this, we apply \SITAcronym\wspace in \proposedModelAcronym\wspace framework with $K \in \{3,5\}$.


\textbf{Comparison with \fedweitAcronym}: Such an approach of parameter selection by assessing domain relevance is missing in baseline \fedweitAcronym\wspace framework, where to control the computational complexity and communication cost, each client transmits task-adaptive parameters of only the previous task to the server.
However, as already discussed, these parameters could be irrelevant for learning current task, thus resulting in \icinterference.

\begin{table*}[h]
	\small
	\begin{center}
		\renewcommand*{\arraystretch}{1.4}
		\begin{tabular}{rl||c:c|c|c||c|c|c||c}
			&  &  & \texttt{Without} & \texttt{With \SITAcronym} & \texttt{With \SITAcronym} & \texttt{Without} & \texttt{With \SITAcronym} & \texttt{With \SITAcronym} &  \\
			&  & \texttt{Baseline} & \texttt{\SITAcronym} & \texttt{($K=3$)} & \texttt{($K=5$)} & \texttt{\SITAcronym} & \texttt{($K=3$)} & \texttt{($K=5$)} &  \\
			\cdashline{3-9}
			& \textbf{Datasets} & \textbf{\fedweitAcronym} & \textbf{\proposedModelAcronym} & \textbf{\proposedModelAcronym} & \textbf{\proposedModelAcronym} & \textbf{\proposedModelAcronym} & \textbf{\proposedModelAcronym} & \textbf{\proposedModelAcronym} & \textbf{Gain} \\
			&  &  &  &  &  & \textbf{+DLS} & \textbf{+DLS} & \textbf{+DLS} & \textbf{(\%)} \\
			\hline
			\hline
			\multirow{5}{*}{\rotatebox{90}{$\lambda_2=1.0$}} & \textbf{R8} & 79.1 \tablepm{2.5} 
			& 90.5 \tablepm{0.6} & 90.5 \tablepm{0.5} & \textbf{90.6 \tablepm{0.7}} & 83.1 \tablepm{5.0} & 82.3 \tablepm{5.9} & 82.4 \tablepm{3.5} & $\uparrow$ 14.5 \\
			\cline{2-10}
			& \textbf{TMN} & 84.2 \tablepm{0.9} 
			& 88.2 \tablepm{0.2} & 88.3 \tablepm{0.1} & \textbf{88.6 \tablepm{0.3}} & 84.7 \tablepm{2.4} & 84.3 \tablepm{1.9} & 83.6 \tablepm{1.3} & $\uparrow$ 5.2 \\
			\cline{2-10}
			& \textbf{TREC6} & 78.2 \tablepm{1.9} 
			& \textbf{83.6 \tablepm{0.6}} & 83.6 \tablepm{0.5} & 83.3 \tablepm{0.2} & 78.2 \tablepm{2.1} & 78.3 \tablepm{1.5} & 77.7 \tablepm{1.8} & $\uparrow$ 6.9 \\
			\cline{2-10}
			& \textbf{TREC50} & 85.1 \tablepm{1.1} 
			& 88.4 \tablepm{2.5} & 88.5 \tablepm{2.4} & \textbf{88.7 \tablepm{1.9}} & 86.4 \tablepm{1.1} & 86.3 \tablepm{1.5} & 85.2 \tablepm{0.2} & $\uparrow$ 4.2 \\
			\cline{2-10}
			& \textbf{SUBJ} & 86.5 \tablepm{0.9} 
			& 88.5 \tablepm{0.6} & 88.6 \tablepm{0.6} & \textbf{89.5 \tablepm{0.3}} & 88.9 \tablepm{0.5} & 89.7 \tablepm{1.7} & 89.0 \tablepm{0.9} & $\uparrow$ 3.5 \\
			\hline
			\hline
			\multirow{5}{*}{\rotatebox{90}{$\lambda_2=0.1$}} & \textbf{R8} & 68.1 \tablepm{10.4} 
			& \textbf{89.9 \tablepm{0.7}} & 89.9 \tablepm{1.1} & 89.7 \tablepm{0.8} & 60.7 \tablepm{14.5} & 61.9 \tablepm{14.9} & 65.3 \tablepm{10.6} & $\uparrow$ 32 \\
			\cline{2-10}
			& \textbf{TMN} & 80.9 \tablepm{0.6} 
			& \textbf{88.5 \tablepm{0.1}} & 88.4 \tablepm{0.07} & 88.3 \tablepm{0.7} & 72.8 \tablepm{9.1} & 73.3 \tablepm{9.8} & 71.1 \tablepm{8.5} & $\uparrow$ 9.4 \\
			\cline{2-10}
			& \textbf{TREC6} & 77.1 \tablepm{3.6} 
			& 84.7 \tablepm{1.1} & \textbf{84.9 \tablepm{1.3}} & 84.7 \tablepm{1.2} & 78.3 \tablepm{1.6} & 78.9 \tablepm{1.3} & 76.8 \tablepm{3.0} & $\uparrow$ 10.1 \\
			\cline{2-10}
			& \textbf{TREC50} & 76.1 \tablepm{0.9} 
			& \textbf{81.2 \tablepm{3.5}} & 81.2 \tablepm{2.9} & 80.9 \tablepm{2.1} & 69.8 \tablepm{3.9} & 69.4 \tablepm{4.7} & 68.9 \tablepm{1.2} & $\uparrow$ 6.7 \\
			\cline{2-10}
			& \textbf{SUBJ} & 85.6 \tablepm{0.6} 
			& 89.1 \tablepm{0.4} & 88.8 \tablepm{0.1} & \textbf{90.4 \tablepm{1.5}} & 88.1 \tablepm{1.3} & 90.2 \tablepm{1.8} & 84.2 \tablepm{2.6} & $\uparrow$ 5.6 \\
		\end{tabular}
	\end{center}
	\caption{Comparison of our proposed \proposedModelAcronym\wspace framework (with and without \SITAcronym) against \fedweitAcronym\wspace baseline model using Task-averaged Test Accuracy (TTA) scores for two different values of $\lambda_2 \in \{1.0, 0.1\}$. Best score for each dataset (row) is shown in \textbf{bold} and \textbf{Gain (\%)} denotes Bold vs \fedweitAcronym.  
	}
	\label{tab:comparison_with_baseline}
\end{table*}

\section{Experiments and Analysis}
To demonstrate the effectiveness of our proposed \withspace{\proposedModelAcronym} framework, we perform evaluation on Text Classification task using five datasets from diverse domains and present our qualitative and quantitative analysis in FCL setup.

\textbf{Datasets:} We present experimental evaluation results on Reuters8 (R8), Tag My News (TMN), TREC6, TREC50 and Subjectivity (SUBJ) datasets. 
Here, R8 and TMN datasets belong to the \textit{News} domain, TREC6 and TREC50 belong to \textit{Question Classification} domain and SUBJ is a \textit{movie reviews} dataset with binary labels. 
Please see Appendix~\ref{sec:datasets_appendix} for more details regarding datasets.

\textbf{Experimental setup:} We follow~\citet{DBLP:conf/icml/YoonJLYH21} for our FCL experimental setup where we use CNN based Text Classification model~\cite{kim-2014-convolutional} as the local client model. 
For all experiments, we use three clients i.e., $C=3$, five tasks per client i.e., $T=5$, 10 rounds per task i.e., $R=10$, with 50 epochs in each round, $\lambda_2 \in \{0.1,1.0\}$ and $K \in \{3,5\}$ for \SITFullName\wspace (\SITAcronym).
Please see Appendix~\ref{sec:exp_settings_appendix} for detailed hyperparameter settings for all of our experiments. 
To run the experiments in FCL setup, we need to generate task datasets $\mathcal{T}_c^t$ for each task $t$. 
Consider a dataset $\dataset_d$ with a unique label set $L_d \equiv \{L_1,...,L_{N_d}\}$, where $N_d$ is the count of unique labels.
Now, for each task dataset $\mathcal{T}_c^t$, we randomly pick a fixed number of unique labels $L_c^t \subseteq L_d$, where the count of unique task labels $|L_c^t| = 4$ is fixed for all tasks across all clients except subjectivity, which has 2 unique labels shared by all tasks.
If label $L_1$ gets selected for 3 tasks, then we follow the \textit{non-iid splitting strategy} which simply divides the documents labeled with $L_1$ i.e., $\dataset_d(L_1)$, into three mutually exclusive and equal parts, thus ensuring heterogeneity.
We use this strategy to split the training and validation sets. 
However, to create the test set for each task dataset $\mathcal{T}_c^t$, we select all of the documents labeled with $L_c^t$ in the complete test dataset i.e., $\{\dataset_d^{test}(L_i) | L_i \in L_c^t\}$ without splitting.
As this work focuses on the new challenges which arise due to combination of FL and CL paradigms in FCL setup, we compare our work with related FCL methods and not with standalone FL, CL methods.

\textbf{Baseline:} 
As the only existing work in FCL domain, \fedweitAcronym\wspace has shown significantly superior performance compared to n\"aive FCL methods that is why we adopt \fedweitAcronym\wspace as the baseline method.

\textbf{Evaluation Metric:}
In each experiment, once the training is finished for all $C$ clients, we freeze the model parameters for all tasks and compute \textit{Micro-averaged Accuracy} (MAA) score for each of the $T$ past tasks of all $C$ clients.
Finally, we average $C \times T$ MAA scores to compute the final Task-averaged Test Accuracy (TTA) score to compare our proposed model with baseline.
For each experiment, we report an average TTA score of 3 runs using 3 different seed values for ordering of tasks at each client.

\begin{figure*}[]
	\centering
	\includegraphics[width=\textwidth]{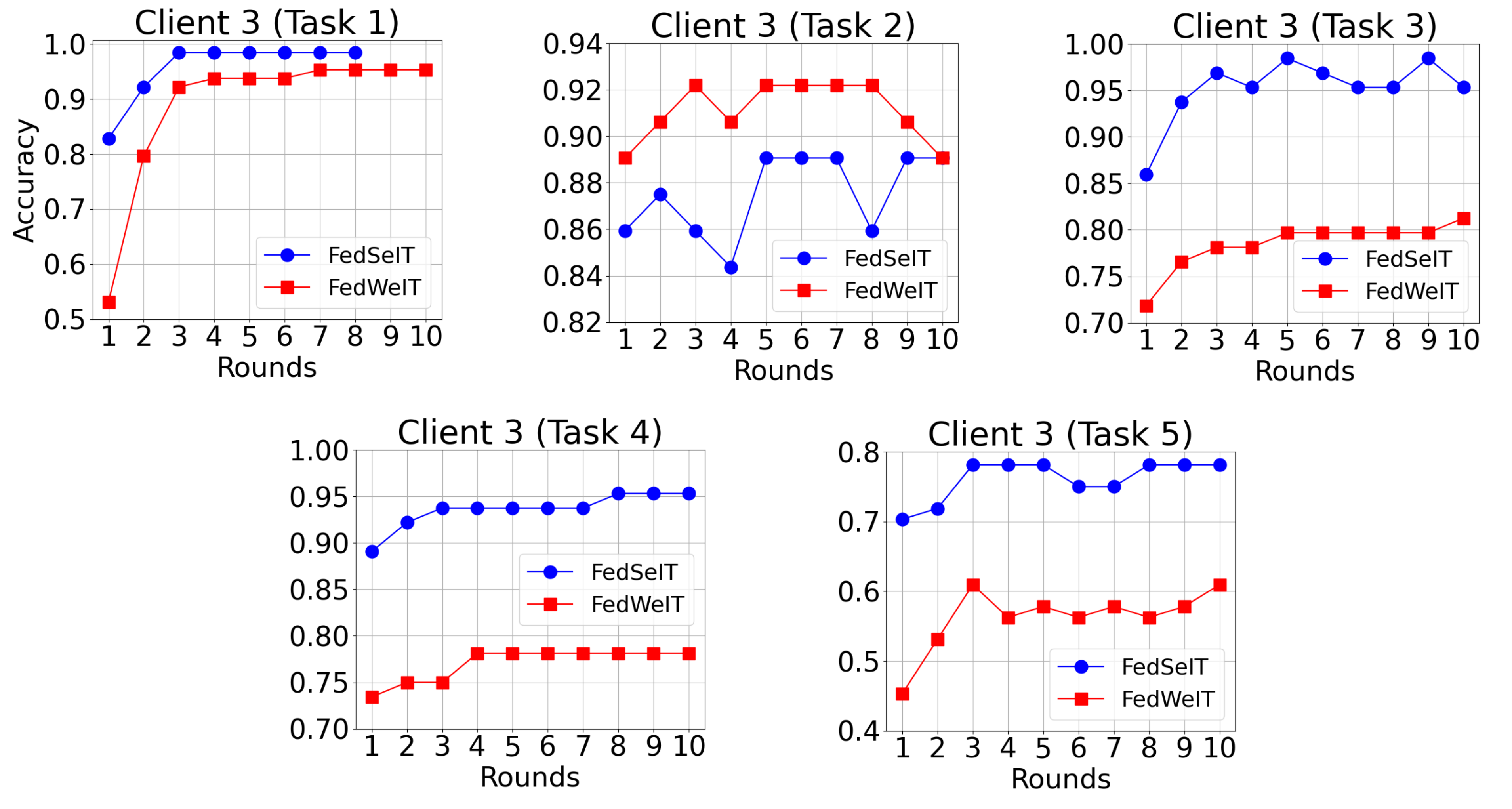}
	\caption{Test set accuracy scores for all 5 tasks of Client-3 in \proposedModelAcronym\wspace framework using Reuters8 dataset ($\lambda_2=1.0$). Each data point denotes the test accuracy score at the end of each training round for 10 rounds of 50 epochs.}
	\label{fig:test_accuracy_plots}
\end{figure*}

\subsection{Results: Comparison to baseline}
Table~\ref{tab:comparison_with_baseline} shows final TTA scores after completion of all tasks across all clients on five Text Classification datasets. 
We find the following observations: 

\textbf{(1)} For all the datasets, our proposed model \proposedModelAcronym\wspace consistently outperforms the baseline method \fedweitAcronym. 
For instance, for the data R8, the classification accuracy is $79.1$\% vs $90.5$\% from \fedweitAcronym\wspace and \proposedModelAcronym\wspace (without SIT), respectively. Overall (column 3 vs column 2), on an average over the five datasets, \proposedModelAcronym\wspace model outperforms  \fedweitAcronym\wspace by 6.4\% and 12.4\% for $\lambda_2$ = $1.0$ (higher penalty on catastrophic forgetting) and  $\lambda_2$ = $0.1$ (lower penalty) respectively. We then observe that \proposedModelAcronym\wspace also applies to sparse-setting, for example, small (R8) vs large  (TMN) corpora. See Table \ref{sec:datasets_appendix} for the data statistics. The results suggest that the selective utilization and domain-alignment of task-adaptive parameters at local clients prevent inter-client interference and maximise transfer learning. 



\textbf{(2)} To demonstrate the effectiveness of \SITFullName\wspace (\SITAcronym) approach by limiting the number of foreign parameters ($K=3$ or $5$) at local client while augment-learning, Table~\ref{tab:comparison_with_baseline} shows a comparison between \proposedModelAcronym\wspace models (trained without and with \SITAcronym) and \fedweitAcronym\wspace baseline.
For all the datasets, it is observed that the test scores of \proposedModelAcronym\wspace trained with \SITAcronym\wspace ($K=3$) are similar to \proposedModelAcronym\wspace trained without \SITAcronym\space ($C=3$) while having the same computational expense; however, reducing the number of parameters when the length of sequence of historical tasks grows at the clients.

Additionally, we observe an improved performance in the classification by increasing $K=5$, i.e., extending the window-size of the number of foreign parameters (considered in augmented-learning) from a sequence of historical tasks in  continual learning. 
For instance, on an average, the performance gain  of \proposedModelAcronym\wspace vs \fedweitAcronym\wspace (column 5 and 2) is:  6.8\% and 12.5\% for $\lambda_2$ = $1.0$ and $\lambda_2$ = $0.1$, respectively. 
This suggests that the more relevant and domain-aligned foreign parameters boost the augmented-learning at each local client, i.e, by selecting relevant foreign parameters from all historical tasks of foreign clients using \SITAcronym. 


\textbf{(3)} Next, we investigate the application of dense-layer-sharing (DLS) in \proposedModelAcronym\wspace, in order to compare with \proposedModelAcronym\wspace (without DLS) and \fedweitAcronym.  For all the datasets, note that the \proposedModelAcronym\wspace outperforms
\proposedModelAcronym+DLS, validating our hypothesis that using exclusive dense layer parameters at local client boots domain-alignment and identification of relevant foreign parameters.  
Overall an average over the five datasets, \proposedModelAcronym\wspace outperforms \proposedModelAcronym+DLS (column 3 vs 6) by 4.4\% and 19.1\% for $\lambda_2$=$1.0$ and $\lambda_2$=$0.1$, respectively.

In summary, evaluating against the baseline \fedweitAcronym\wspace approach,
the proposed \proposedModelAcronym\wspace have shown improved performance, a average gain of \textbf{12.4\%} in text classification over a sequence of tasks using the five datasets from diverse domains.

\subsection{Ablation: Learning over training rounds}
Here, we demonstrate the performance of text classification at a local client (3rd client) for a sequence of five tasks at each training round (10 rounds). 
Figure~\ref{fig:test_accuracy_plots} shows the test set accuracy scores (at the end of each training round) for all five tasks of Client-3 using R8 dataset. 
In 4/5 tasks, the test accuracy score of the \proposedModelAcronym\wspace model at the end of round 1 is noticeably higher than the \fedweitAcronym approach.
Interestingly, in task 2, we find that \fedweitAcronym\wspace outperforms \proposedModelAcronym\wspace over rounds, however converge at the same accuracy in the final round. 

In essence, the proposed method \proposedModelAcronym\wspace in FCL setup have shown that the alignment and relevance of foreign tasks parameters at each client (for all the tasks at each model training round) minimise inter-client interference and improve inter-client transfer learning without the dense-layer-sharing.  

\section{Conclusion and Future Work}
We have applied Federated Continual Learning to text classification for heterogeneous tasks and addressed the challenges of inter-client interference and domain-alignment in model parameters of local vs foreign clients while minimizing catastrophic forgetting over a sequence of tasks. We have presented two novel techniques:  \proposedModelAcronym\wspace and \SITAcronym\wspace that improves local client augmented-learning by assessing domain overlap and selecting informative tasks from the sequence of historical tasks of each foreign client while preserving privacy. Furthermore, the novel selection strategy using SIT determines relevant foreign tasks from the complete historical tasks of all foreign clients by assessing domain overlap while preserving privacy. We have evaluated the proposed approaches using five text classification data sets and shown a gain (average) of 12.4\% over the baseline.

Although we have applied \proposedModelAcronym\wspace framework to the document-level text classification task, we can further apply the proposed framework to additional NLP tasks.
Inspired by continual Topic Modeling~\cite{gupta2020neural} at document-level and continual Named Entity Recognition~\cite{cl_ner} at token-level classification, we can further extend these existing works with the proposed \proposedModelAcronym\wspace framework in federated settings.

\section*{Limitations}
In FCL paradigm, the computation complexity of augmented-learning at a client increases when the number of foreign clients grows exponentially. In future work, it is an interesting research direction to explore hierarchical federated learning techniques \cite{hierarchicalfederatedlearning} to limit the number of foreign client parameters injected into augmented-learning (applying convolution filters and projections in CNN of \proposedModelAcronym) at a local client. 
Additionally, due to limited compute on edge devices such as mobiles, wearable devices, sensors, etc., the application of FCL in the cloud-edge continuum is still in early days that requires distillation and pruning of large ML models such as CNN for text classification - as presented in this paper. 

\section*{Acknowledgements}

This research was supported (also executed at) by DRIMCo GmbH, Munich, Germany.
We are also grateful to anonymous reviewers for their constructive comments and suggestions.

\bibliography{anthology,custom}

\begin{thebibliography}{18}
\expandafter\ifx\csname natexlab\endcsname\relax\def\natexlab#1{#1}\fi

\bibitem[{Abad et~al.(2020)Abad, Ozfatura, Gunduz, and
  Ercetin}]{hierarchicalfederatedlearning}
Mehdi Salehi~Heydar Abad, Emre Ozfatura, Deniz Gunduz, and Ozgur Ercetin. 2020.
\newblock Hierarchical federated learning across heterogeneous cellular
  networks.
\newblock In \emph{ICASSP 2020-2020 IEEE International Conference on Acoustics,
  Speech and Signal Processing (ICASSP)}, pages 8866--8870. IEEE.

\bibitem[{Caldas et~al.(2018)Caldas, Smith, and
  Talwalkar}]{caldas2018federated}
Sebastian Caldas, Virginia Smith, and Ameet Talwalkar. 2018.
\newblock Federated kernelized multi-task learning.
\newblock In \emph{Proc. SysML Conf.}, pages 1--3.

\bibitem[{Gupta et~al.(2020)Gupta, Chaudhary, Runkler, and
  Schuetze}]{gupta2020neural}
Pankaj Gupta, Yatin Chaudhary, Thomas Runkler, and Hinrich Schuetze. 2020.
\newblock Neural topic modeling with continual lifelong learning.
\newblock In \emph{International Conference on Machine Learning}, pages
  3907--3917. PMLR.

\bibitem[{Karimireddy et~al.(2020)Karimireddy, Kale, Mohri, Reddi, Stich, and
  Suresh}]{karimireddy2020scaffold}
Sai~Praneeth Karimireddy, Satyen Kale, Mehryar Mohri, Sashank Reddi, Sebastian
  Stich, and Ananda~Theertha Suresh. 2020.
\newblock Scaffold: Stochastic controlled averaging for federated learning.
\newblock In \emph{International Conference on Machine Learning}, pages
  5132--5143. PMLR.

\bibitem[{Kim(2014)}]{kim-2014-convolutional}
Yoon Kim. 2014.
\newblock Convolutional neural networks for sentence classification.
\newblock In \emph{Proceedings of the 2014 Conference on Empirical Methods in
  Natural Language Processing ({EMNLP})}, pages 1746--1751.

\bibitem[{Kirkpatrick et~al.(2017)Kirkpatrick, Pascanu, Rabinowitz, Veness,
  Desjardins, Rusu, Milan, Quan, Ramalho, Grabska-Barwinska
  et~al.}]{kirkpatrick2017overcoming}
James Kirkpatrick, Razvan Pascanu, Neil Rabinowitz, Joel Veness, Guillaume
  Desjardins, Andrei~A Rusu, Kieran Milan, John Quan, Tiago Ramalho, Agnieszka
  Grabska-Barwinska, et~al. 2017.
\newblock Overcoming catastrophic forgetting in neural networks.
\newblock \emph{Proceedings of the national academy of sciences}, 114.

\bibitem[{Kumar and Daume~III(2012)}]{kumar2012learning}
Abhishek Kumar and Hal Daume~III. 2012.
\newblock Learning task grouping and overlap in multi-task learning.
\newblock \emph{arXiv preprint arXiv:1206.6417}.

\bibitem[{Li et~al.(2020)Li, Sahu, Zaheer, Sanjabi, Talwalkar, and
  Smith}]{li2020federated}
Tian Li, Anit~Kumar Sahu, Manzil Zaheer, Maziar Sanjabi, Ameet Talwalkar, and
  Virginia Smith. 2020.
\newblock Federated optimization in heterogeneous networks.
\newblock \emph{Proceedings of Machine Learning and Systems}, 2:429--450.

\bibitem[{McMahan et~al.(2017)McMahan, Moore, Ramage, Hampson, and
  y~Arcas}]{fedavg}
Brendan McMahan, Eider Moore, Daniel Ramage, Seth Hampson, and Blaise~Aguera
  y~Arcas. 2017.
\newblock Communication-efficient learning of deep networks from decentralized
  data.
\newblock In \emph{Artificial intelligence and statistics}, pages 1273--1282.
  PMLR.

\bibitem[{Mikolov et~al.(2013)Mikolov, Chen, Corrado, and
  Dean}]{mikolov2013efficient}
Tomas Mikolov, Kai Chen, Greg Corrado, and Jeffrey Dean. 2013.
\newblock Efficient estimation of word representations in vector space.
\newblock \emph{arXiv preprint arXiv:1301.3781}.

\bibitem[{Monaikul et~al.(2021)Monaikul, Castellucci, Filice, and
  Rokhlenko}]{cl_ner}
Natawut Monaikul, Giuseppe Castellucci, Simone Filice, and Oleg Rokhlenko.
  2021.
\newblock Continual learning for named entity recognition.
\newblock \emph{Proceedings of the AAAI Conference on Artificial Intelligence},
  pages 13570--13577.

\bibitem[{Schwarz et~al.(2018)Schwarz, Czarnecki, Luketina, Grabska-Barwinska,
  Teh, Pascanu, and Hadsell}]{schwarz2018progress}
Jonathan Schwarz, Wojciech Czarnecki, Jelena Luketina, Agnieszka
  Grabska-Barwinska, Yee~Whye Teh, Razvan Pascanu, and Raia Hadsell. 2018.
\newblock Progress \& compress: A scalable framework for continual learning.
\newblock In \emph{International Conference on Machine Learning}, pages
  4528--4537. PMLR.

\bibitem[{Shoham et~al.(2019)Shoham, Avidor, Keren, Israel, Benditkis,
  Mor{-}Yosef, and Zeitak}]{fedcurv}
Neta Shoham, Tomer Avidor, Aviv Keren, Nadav Israel, Daniel Benditkis, Liron
  Mor{-}Yosef, and Itai Zeitak. 2019.
\newblock Overcoming forgetting in federated learning on non-iid data.
\newblock \emph{CoRR}, abs/1910.07796.

\bibitem[{Thrun(1995)}]{thrun1995lifelong}
Sebastian Thrun. 1995.
\newblock A lifelong learning perspective for mobile robot control.
\newblock In \emph{Intelligent robots and systems}, pages 201--214. Elsevier.

\bibitem[{Wang et~al.(2019)Wang, Han, Wang, Zhao, Chen, and
  Chen}]{edgecomputing}
Xiaofei Wang, Yiwen Han, Chenyang Wang, Qiyang Zhao, Xu~Chen, and Min Chen.
  2019.
\newblock In-edge ai: Intelligentizing mobile edge computing, caching and
  communication by federated learning.
\newblock \emph{IEEE Network}.

\bibitem[{Yoon et~al.(2021)Yoon, Jeong, Lee, Yang, and
  Hwang}]{DBLP:conf/icml/YoonJLYH21}
Jaehong Yoon, Wonyong Jeong, Giwoong Lee, Eunho Yang, and Sung~Ju Hwang. 2021.
\newblock Federated continual learning with weighted inter-client transfer.
\newblock In \emph{{ICML}}, pages 12073--12086. {PMLR}.

\bibitem[{Yurochkin et~al.(2019)Yurochkin, Agarwal, Ghosh, Greenewald, Hoang,
  and Khazaeni}]{yurochkin2019bayesian}
Mikhail Yurochkin, Mayank Agarwal, Soumya Ghosh, Kristjan Greenewald, Nghia
  Hoang, and Yasaman Khazaeni. 2019.
\newblock Bayesian nonparametric federated learning of neural networks.
\newblock In \emph{International Conference on Machine Learning}, pages
  7252--7261. PMLR.

\bibitem[{Zhang et~al.(2020)Zhang, Kuang, You, Shen, Xiao, Zhang, Wu, Zhuang,
  and Li}]{zhang2020federated}
Fengda Zhang, Kun Kuang, Zhaoyang You, Tao Shen, Jun Xiao, Yin Zhang, Chao Wu,
  Yueting Zhuang, and Xiaolin Li. 2020.
\newblock Federated unsupervised representation learning.
\newblock \emph{arXiv preprint arXiv:2010.08982}.

\end{thebibliography}
\bibliographystyle{acl_natbib}

\appendix

\section{Datasets}
\label{sec:datasets_appendix}

Table~\ref{table:datastatistics} shows detailed data statistics of five labeled datasets used to evaluate our proposed \proposedModelAcronym\wspace framework using Text Classification task: Reuters8 (R8) and Tag My News (TMN) datasets belong to the \textit{News} domain, TREC6 and TREC50 belong to the \textit{Question Classification} domain and Subjectivity (SUBJ) is a \textit{movie reviews} dataset with binary labels. 

\begin{table*}[h!]
	\begin{center}
		\label{r8_stats}
		\begin{tabular}{r|c|c|c|c|c|c}
	  & \textbf{\#Train} & \textbf{\#Test} & \textbf{Maximum} & \textbf{Mean} & \textbf{Median} & \textbf{Num} \\
			\textbf{Datasets} & \textbf{docs} & \textbf{docs} & \textbf{length*} & \textbf{length*} & \textbf{length*} & \textbf{Classes} \\
			\hline
			\textbf{Reuters8} & 5485 & 2189 & 964 & 102 & 64 & 8  \\
			\textbf{TMN} & 24816 & 7779 & 48 & 18 & 18 & 7  \\
			\textbf{TREC6} & 5451 & 499 & 37 & 10 & 9 & 6  \\
			\textbf{TREC50} & 5451 & 499 & 37 & 10 & 9 & 50  \\
			\textbf{SUBJ} & 7999 & 1999 & 120 & 24 & 22 & 2  \\
		\end{tabular}
	\end{center}
	\caption{Detailed statistics of the datasets used for evaluation of our proposed \proposedModelAcronym\wspace framework on Text Classification task. Here, * indicates that length is equal to the count of word tokens.}
	\label{table:datastatistics}
\end{table*}

\begin{table*}[h!]
	\begin{center}
		\begin{tabular}{p{5cm}|c|p{7.5cm}}
			\hline
			\textbf{Hyperparameter} & \textbf{Value} & \textbf{Description}\\
			\hline
			$\lambda_2$ & \{0.1, 1.0\} & Higher value minimizes catastrophic forgetting \\
			$K$ & \{3, 5\} & Number of tasks sent to each client from server after assessing domain overlap \\
			\hline
			$L_{Conv}$ & 3 & Number of convolution layers \\
			$\mathcal{F}$ & \{3, 4, 5\} & Kernel Sizes of convolution filters \\
			$\mathcal{N}^{\mathcal{F}}$ & 128 & Number of convolution filters in each convolution layer \\
			Stride & 1 & Stride when applying convolution filter \\
			Dilation Rate & 1 & Dilation rate when applying convolution filter \\
			Padding & Valid & Padding when applying convolution filter \\
			\hline
			Dropout & 0.3 & - \\
			Learning rate & 1e-4 & - \\
			Activation Function & ReLU & - \\
			Batch Size & 64 & - \\
			$\lambda_1$ & 1e-3 & Regularization parameter for sparsity constraint \\
			\hline
			Random Seeds (for task allocation and parameter initialization) & \{1, 2\} & - \\
			$C$ & 3 & Number of clients \\
			$T$ & 5 & Tasks picked per client \\
			$R$ & 10 & Number of Rounds \\
			Number of epochs per training round & 50 & Early stopping due to convergence is possible \\
			Early stopping patience & 3 & - \\
			\hline
			Random Seeds (for task generation) & 42 & - \\
			Clustering Algorithm for \SITFullName\wspace (\SITAcronym) & KNN & - \\
			$Q$ & 200 & Count of cluster centers to extract \\
			$\mathcal{D}$ & 300 & Word embedding dimension \\
			\hline
		\end{tabular}
	\end{center}
	\caption{Hyperparameter settings used in the experimental evaluation setup of our proposed \proposedModelAcronym\wspace framework using Text Classification task. 
	}
	\label{table:hyperparameters}
\end{table*}

\section{Local client model}
\label{sec:client_model_appendix}

In \proposedModelAcronym\wspace we use CNN~\cite{kim-2014-convolutional} model for Text Classification as the local client model. 
The CNN model is made up of convolution (Conv) layers and fully connected (FC) layers.
The CNN model parameters can be described as follows: 
(1) For Conv layers: $\B_c^t, \A_c^t, \thetaparam_c^t \in \filterparamdim$ are a set of convolutional filters where, $l$ is the layer indicator, $L_{Conv}$ is the total number of Conv layers, $\mathcal{F}_l$ is the filter size for layer $l$, $\mathcal{D}$ is the input word embedding dimension and $\mathcal{N}_l^\mathcal{F}$ is the count of filters in layer $l$, and
(2) For FC layers: $\B_c^t, \A_c^t, \thetaparam_c^t \in \matrixparamdim$ are a set of parameter matrices where, $l$ is the layer indicator, $L_{FC}$ is the total number of FC layers and $\mathcal{I}_l$, $\mathcal{O}_l$ are the input, output dimensions for layer $l$.
For all layers, $\m_c^t$ is the masking vector matching the output dimension of $\B_c^t$.
As illustrated in Figure~\ref{fig:cnn}, for an input document $x$, the CNN model performs three different tasks: 
(1) performing word embedding lookup of $x$ to generate an input matrix $\X \in \mathbb{R}^{|x| \times \mathcal{D}}$, 
(2) applying convolutional filters and max-pooling over \X\wspace to generate an intermediate dense vector $\z \in \mathbb{R}^{\mathcal{N}^\mathcal{F}}$, and
(3) applying softmax layer on \z\wspace to predict the label for input $x$.

\section{Experimental Setup}
\label{sec:exp_settings_appendix}

Table~\ref{table:hyperparameters} shows the settings of all hyperparameters used in the experimental setup to evaluate our proposed \proposedModelAcronym\wspace framework on Text Classification task using 5 datasets.

\section{Reproducibility: Code}
\label{sec:reproducibility}

To run the experiments and reproduce the scores reported in the paper content, we have provided the implementation of our proposed \proposedModelAcronym\wspace framework at \url{https://github.com/RaiPranav/FCL-FedSeIT}.
Information regarding model training and data pre-processing is provided in the README file.

\end{document}